\pdfoutput=1

\documentclass[11pt]{article}

\usepackage[]{emnlp2021}
\usepackage{times}
\usepackage{latexsym}
\usepackage{multirow}
\usepackage{amsmath}
\usepackage{cleveref}
\usepackage{algorithm}
\usepackage{algorithmicx}
\usepackage{algpseudocode}
\usepackage{graphicx}
\usepackage{url}
\usepackage{graphics}

\usepackage{latexsym}

\crefname{section}{§}{§§}
\usepackage[T1]{fontenc}

\usepackage[utf8]{inputenc}

\usepackage{microtype}

\title{AR-LSAT: Investigating Analytical Reasoning of Text}

\author{Wanjun Zhong$^1$\thanks{\ \ \ Work done while this author was an intern at Microsoft Research.} , Siyuan Wang$^3$$^*$, Duyu Tang$^2$, Zenan Xu$^1$$^*$, Daya Guo$^1$$^*$\\
	\bf Yining Chen$^2$, Jiahai Wang$^1$, Jian Yin$^1$, Ming Zhou$^4$ and Nan Duan$^2$ \\
	$^1$ The School of Data and Computer Science, Sun Yat-sen University.\\
	$^2$ Microsoft Research  $^3$ Fudan University, China $^4$SINOVATION VENTURES\\
	{\tt \{zhongwj25, xuzn, guody5\}@mail2.sysu.edu.cn}\\
	{\tt \{wangjiah@mail,issjyin@mail\}.sysu.edu.cn} \\
	{\tt \{dutang,nanduan,yining.chen\}@microsoft.com}\\ 
	{\tt wangsy18@fudan.edu.cn}; {\tt zhouming@chuangxin.com} \\
}

\begin{document}
\maketitle
\begin{abstract}
Analytical reasoning is an essential and challenging task that requires a system to analyze a scenario involving a set of particular circumstances and perform reasoning over it to make conclusions.
In this paper, we study the challenge of analytical reasoning of text and introduce a new dataset consisting of questions from the Law School Admission Test from 1991 to 2016.
We analyze what knowledge understanding and reasoning abilities are required to do well on this task.
Furthermore, to address this reasoning challenge, we design two different baselines: 
(1) a Transformer-based method which leverages the state-of-the-art pre-trained language models
and (2) Analytical Reasoning Machine (ARM), a logical-level reasoning framework extracting symbolic knowledge (e.g, participants, facts, logical functions) to deduce legitimate solutions.
In our experiments, we find that the Transformer-based models 
struggle to solve this task as their performance is close to random guess and ARM achieves better performance by leveraging symbolic knowledge and interpretable reasoning steps. 
Results show that both methods still lag far behind human performance, which leave further space for future research. 
\footnote{The data and code are provided in \url{https://github.com/zhongwanjun/AR-LSAT}.}

\end{abstract}
\begin{figure}[t]
     \includegraphics[width=0.42\textwidth]{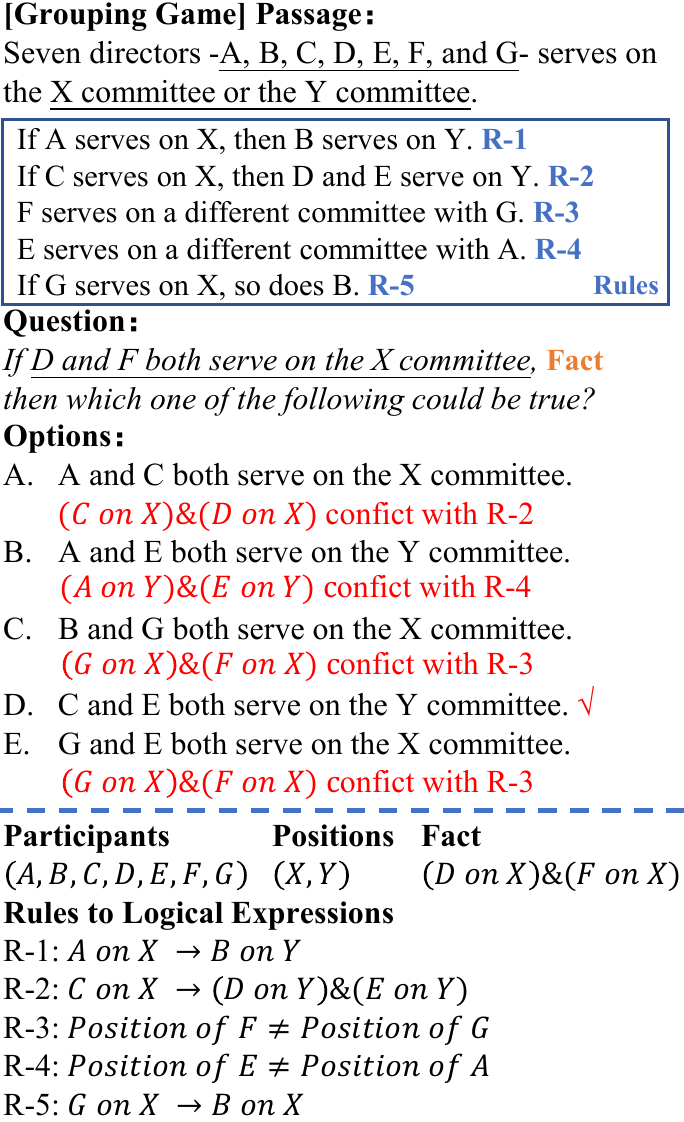}
     \caption{An example of the required reasoning process to do well on the AR task. The input is a passage, a question and multiple options, and the output is the most plausible answer.}
     \label{fig:first-example}
\end{figure}

\section{Introduction}

Analytical reasoning assesses the problem-solving ability to understand knowledge (e.g., participants, facts, rules), and reasoning over that knowledge to determine a solution.
Analytical reasoning is known to involved when doing everyday tasks, and engages high-level cognitive mechanisms of humans \cite{williams2019thinking}. 
Although Transformer-based pre-trained language models including BERT \cite{devlin2018bert}, GPT-2 \cite{radford2019language} and RoBERTa \cite{liu2019roberta} 
have achieved state-of-the-art performance on a variety of NLP tasks, they still struggle to perform deep reasoning beyond shallow-level semantic understanding of literal clues.
For example, \newcite{talmor2020olmpics} show that pre-trained models fail on half of eight
reasoning tasks that require symbolic operations. 
We hope to challenge current systems and take a step towards analytical reasoning.

In this paper, we study the challenge of analytical reasoning (AR). 
We introduce a new dataset \textbf{AR-LSAT} from the Law School Admission Test\footnote{\url{https://en.wikipedia.org/wiki/Law_School_ Admission_Test}} (LSAT) from 1991 to 2016.
to facilitate research on this area. 
An example of analytical reasoning in LSAT is given in Figure \ref{fig:first-example}, whose task is to separate participants (i.e., \textit{A,B, etc.}) into two positions (i.e., \textit{X committee and Y committee}) under certain constraints.
Solving the problem requires a system to understand the knowledge in the context including participants, positions, rules expressed in natural language (e.g., ``\textit{If G serves on X, so does B}") and facts (e.g., ``\textit{D and F both serve on the X committee}"). 
Then, it needs to deduct logical expressions (e.g., ``$\text{\textit{G on X}}\rightarrow \text{\textit{B on X}}$") from the rules, and draw inference
before making conclusions. 

In this paper, we analyze the knowledge understanding and reasoning ability required for solving this task and present two base approaches for this challenge: 
(1) Transformer-based approach that applies pretrained language models to encode the input context into distributed representation for classification. 
(2) Analytical Reasoning Machine (ARM), a logical-level framework that first extracts symbolic knowledge (i.e., participants, rules, facts) from the context, and further maps them into executable logical functions (e.g., ``\textit{IfThen}", ``\textit{Before}") to assess whether a solution can satisfy mentioned rules and then deduce legitimate solutions for making prediction. 
This framework sheds a light on the logical-level reasoning procedure required for this task, and each step can be further developed in future for better performance or expandability.

\begin{figure*}[thbp]
     \centering
     \includegraphics[width=\textwidth]{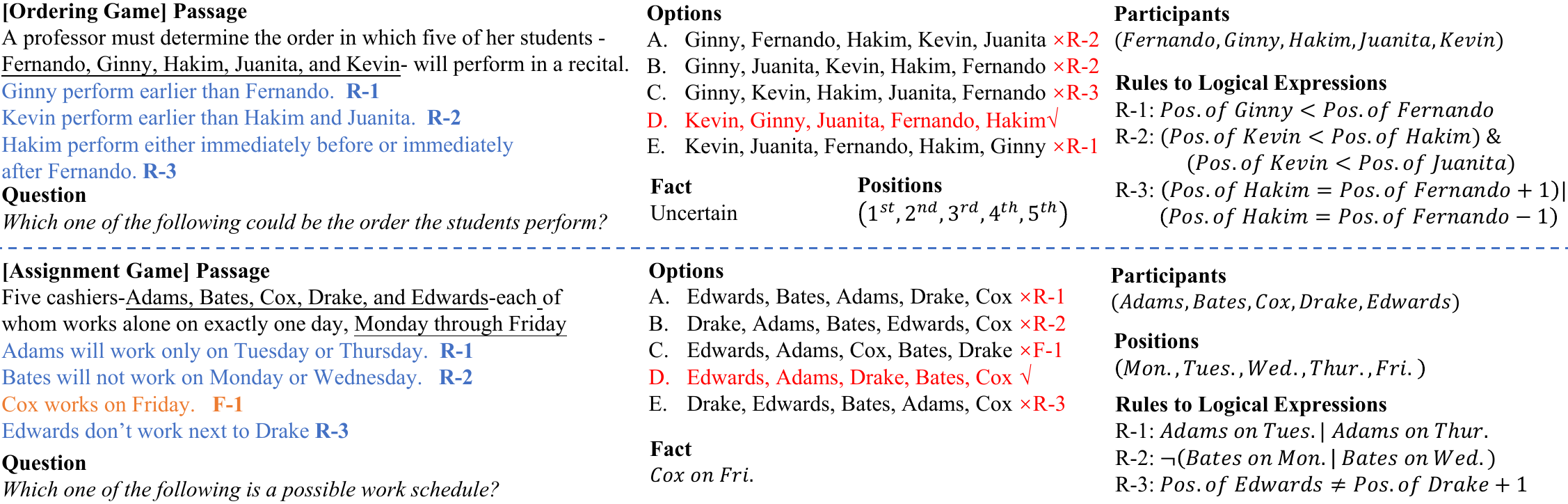}
     \caption{Examples of ordering game and assignment game in AR task. Facts and Rules are highlighted in orange and blue, respectively. Example of grouping game is shown in Figure \ref{fig:first-example}. $\times$ indicates conflict.}
     \label{fig:examples}
\end{figure*}
Experiments show that the Transformer-based approach struggles to learn this task, which indicates that this task is very challenging for current models as it requires the complex reasoning ability far beyond implicit reasoning over the literal clues.
ARM performs relatively better than the Transformer-based approach with higher accuracy and better interpretability. 
The performance of both approaches lag far behind human performance, which leaves a huge space for further research.

The contributions of our paper are two-fold.
\begin{itemize}
    \item We introduce a new dataset AR-LSAT to facilitate research on analytical reasoning.
    \item We present two approaches for this task: a Transformer-based approach and a logical-level reasoning framework that utilizes symbolic knowledge to perform reasoning. 
    
\end{itemize}
\section{Related Works}
There is an increasing trend on machine reasoning research in recent years. 
The reasoning ability investigated are partitioned into several major aspects, including (1) logical reasoning; (2) commonsense reasoning; (3) mathematical reasoning and (4) multi-hop reasoning.
\paragraph{Logical Reasoning}
The task of Natural Language Inference (NLI) \cite{inproceedings, snli:emnlp2015, DBLP:journals/corr/abs-1804-07461,williams-etal-2018-broad, welleck2018dialogue, scitail,Nie2019AdversarialNA, ch2019abductive,liu2020natural} requires the models to detect the logical entailment relationship of two sentences. 
There have been Machine Reading Comprehension (MRC) datasets \cite{rajpurkar2016squad, welbl2017constructing, Yang_2018,cosmos} that examine the ability of logical reasoning. 
LogiQA \cite{Liu_2020} and ReClor \cite{yu2020reclor} are sourced from examination in realistic scenario and examine a range of logical reasoning skills.
\paragraph{Commonsense Reasoning}
There are many recent benchmarks that assess the commonsense reasoning capabilities from different aspects, like social \cite{rashkin2018event2mind}, physics \cite{talmor2018commonsenseqa, zellers2019hellaswag}, or temporal \cite{zhou2019going}.
There exist several MRC datasets that require commonsense knowledge \cite{ostermann2018semeval, zhang2018record,huang2019cosmos}.

\paragraph{Mathematical Reasoning}
There are many existing datasets \cite{kushman2014learning, hosseini2014learning, koncel2015parsing,clark2016combining,ling2017program} focus on mathematical word problems. 
\citet{ling2017program} builds a dataset that encourages generating answer rationales beyond simply selecting the correct answer.
DROP \cite{dua2019drop} is a benchmark MRC dataset requiring mathematical reasoning. \citet{saxton2019analysing} focuses on algebraic generalization. 
\paragraph{Multi-hop Reasoning}
Multi-hop reasoning over textual data \cite{Talmor2018TheWA, welbl2018constructing, yang2018hotpotqa,inoue-etal-2020-r4c} require a model to reason over multiple paragraphs before making prediction. 

To the best of our knowledge, there has not an existing benchmark dataset that completely focuses on the analytical reasoning over textual data. We introduce a new dataset to fill this gap and to foster research on this area.

\section{Task and Dataset}
In this section, we describe the task of analytical reasoning, introduce the dataset AR-LSAT we collected from the Law School Admission Test and make analysis about the required reasoning skills.
\subsection{Task: Analytical Reasoning of Text}
Taking a passage, a question, and multiple options as the input, a system is required to select the most plausible answer as the output.
Each passage describes a reasoning game belonging to various types. According to \citet{LSAT-book-2}, there are three dominant game types in LSAT: \textbf{ordering games}, \textbf{grouping games}, and \textbf{assignment games}, which are described as follows and examples are given in Figures \ref{fig:first-example} and \ref{fig:examples}:
\begin{itemize}
    \item \textbf{Ordering games} are to order participants based on given facts and rules.
    \item \textbf{Grouping games} are to separate participants into groups with given facts and rules. 
    \item \textbf{Assignment games} are to assign characteristics to the participants with given rules, like assigning schedules for people. 
\end{itemize}
\subsection{Dataset Collection: AR-LSAT}
We collect data from nearly 90 LSAT exams from 1991 to 2016 and select questions from the analytical reasoning part to construct the dataset, and name it \textbf{AR-LSAT}. 
Each exam in LSAT consists of 101 multiple choice questions, 24 of which are AR questions.
We finally leave up the questions with 5 answer options. 
\begin{table}[h]
	\centering
	\begin{tabular}{ll}
		\hline
		Number of questions & 2,046 \\
		Average length of passages & 99.3 \\
		Average length of questions & 19.1 \\
		Average length of answers & 6 \\
		Number of options & 5 \\ 
		Ratio of ordering game & 42.5\% \\
		Ratio of grouping game & 38.75\% \\
		Ratio of assignment game & 18.75\% \\
		\hline
	\end{tabular}
\caption{Data statistics of AR-LSAT dataset.}
\label{tab:data-statistic}
\end{table}

\begin{table*}[ht]
\small
\begin{tabular}{ll}
\hline
\textbf{Question Type} & \textbf{Description} \\ \hline
Acceptable solution (15.6\%)& identify a feasible solution that can satisfy all the rules \\
Complete list (3.5\%) & identify a complete and accurate list of participants under given condition \\
Could be true/false (26.8\%) & select answer that could be true/false under given condition \\
Must be true/false (26.4\%) & select answer that must be true/false under given condition \\
Negation (14.7\%) & questions that contain negation \\
Substitution (4.3\%) & identify a new rule that can substitute one of the old rules for the desiring result \\
Condition for determined solution (3.5\%) & identify a new rule so that the feasible solution is determined \\
Calculation (3\%) & calculate possible participants in a group \\
Earliest/latest position (1.3\%) & identify the earliest/latest position that a specific participant can be assigned to \\
Maximum/minimum members (1.3\%)& identify the possible maximum/minimum number of participants in a specific group \\ \hline
\end{tabular}
\caption{The ratio and description of each question type in the test set of the AR-LSAT dataset.}
\label{tab:question-type}
\end{table*}

\subsection{Data Analysis}
As mentioned above, the questions of AR-LSAT come from exams in realistic scenario. 
Each passage describes a reasoning game belongs to three dominant type: (1) ordering game, (2) grouping game and (3) assignment game. 
We manually analyze and summarize the ratio of each type of reasoning game in AR-LSAT. 
The corresponding data statistics and ratios are shown in Table \ref{tab:data-statistic}.
  Moreover, the questions in AR-LSAT are further challenging as them require the system to have different kinds of reasoning skills. We manually categorize and analyze question types that are common in AR-LSAT dataset. The detailed description of question types is shown in Table \ref{tab:question-type}. We also notice that the three most common question types: ``acceptable solution", ``could be true/false" and ``must be true/false" associate with most of the passages. 
  There also exist challenging questions, like ``calculation" and ``substitution" problems.
  The examples of question types are  given in Appendix C.

\subsection{Challenges}
In this part, we point out the reasoning ability required for solving AR questions, and put forward the challenges that systems should face. 
As we can observe from the examples in Figure \ref{fig:first-example} and Figure \ref{fig:examples}, solving AR questions needs systems to understand the complex scenario and perform reasoning over it, and has no special needs for external knowledge. 
In conclusion, AR questions test a range of reasoning skills:
\begin{itemize}
    \item[1)] Comprehending the knowledge including participants of events, facts, and rules described in the context. 
    \item[2)] Extracting machine-understandable logical functions (expressions) from the rules. For example, the rule ``\textit{If A serves on X, then B serves on Y}." needs to be transferred as logical expression ``$\text{\textit{A on X}} \rightarrow \text{\textit{B on Y}}$", 
    \item[3)] Making deductions to derive legitimate solutions that satisfy extracted logical functions. 
    \item[4)] Selecting the answer that satisfies all the rules with the deducted legitimate solutions. In the examples, a system should eliminate options that conflict with rules and select the option that accords with legitimate solutions.
\end{itemize}
Therefore, this task requires the machine to perform explicit complex reasoning, far beyond just understanding the literal clues presented in the text.

\section{Approaches}
In this section, we describe our two base approaches: (1) Transformer-based approach and (2) Analytical Reasoning Machine (ARM).
\subsection{Transformer-based Approach}
In this approach, we view the analytical reasoning challenge as a multiple-choice question answering problem. 
We employ state-of-the-art pre-trained Transformer-based language models (i.e., BERT \cite{devlin2018bert}, XLNet \cite{yang2019xlnet}, RoBERTa \cite{liu2019roberta}, and ALBERT \cite{lan2019albert}) for classification as they achieve impressive performance on a wide variety of tasks.
Specifically, we take the concatenated sequence $X = \{[CLS], passage, [SEP],question,option\}$ as the input, where $[CLS]$ is the ending special token and $[SEP]$ is used to split two types of input.
The representation of the sequence $H=f_{Transformer}(X)$ is further fed into a two-layer perceptron $f_{MLP}$ for classification $p_{\theta}(X)=\sigma(f_{MLP}(H))$, where $\sigma$ is an activation function. The model parameters $\theta$ of the Transformer and MLP layer are fine-tuned with cross-entropy loss on the training set.  

\subsection{Analytical Reasoning Machine (ARM)}
\begin{figure*}[thbp]
     \centering
     \includegraphics[width=\textwidth]{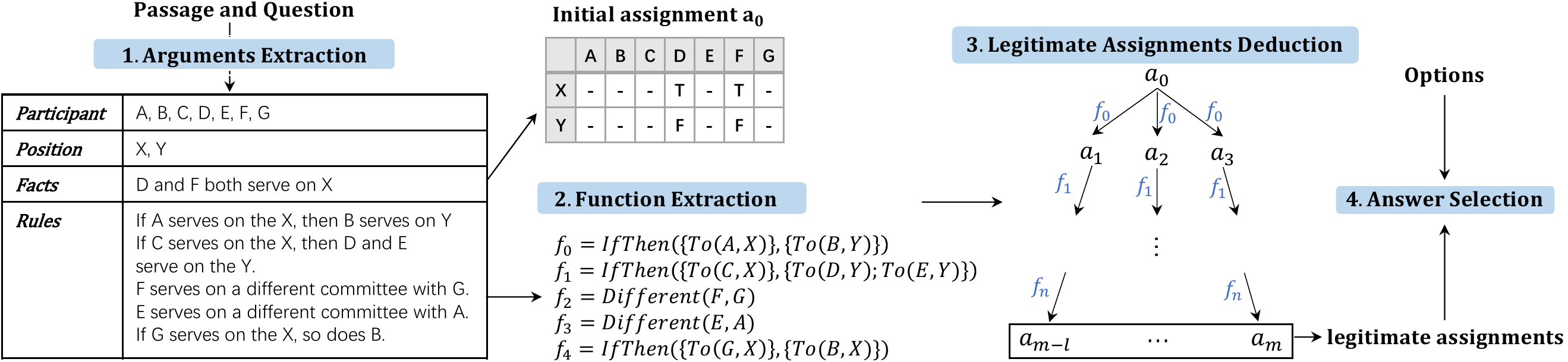}
     \caption{An overview of our approach. The original example is given in Figure \ref{fig:first-example}. It extracts arguments from the context (\cref{sec:scenario-understanding}). Then it extracts logical functions from rules (\cref{sec:function-generation}). Afterwards, it conducts deduction to find legitimate assignments (\cref{sec:reasoning}). Lastly, it matches the options and legitimate assignments for prediction (\cref{sec:answer-selection}).}
     \label{fig:approach}
\end{figure*}
In this part, we describe the logical-level framework, Analytical Reasoning Machine (ARM), which extracts symbolic knowledge from the context and perform reasoning over the knowledge to draw conclusions. 
Figure \ref{fig:approach} gives an overview of the ARM framework.
We propose to break down the reasoning process into four stages:
(1) extracting arguments (i.e., the participants, positions, facts and rules) from the context (\cref{sec:scenario-understanding}); 
(2) 
interpreting rules into a set of logical constraint functions, whose arguments are selected from participants and positions (\cref{sec:function-generation});
(3) reasoning with the logical functions
and finally generating a group of legitimate assignments (solutions) that satisfy all the rules (\cref{sec:reasoning});
(4) selecting the most plausible option by matching the legitimate assignments and options (\cref{sec:answer-selection}).

ARM sheds a light on the logical-level reasoning procedure for analytical reasoning and each procedure can be further developed for both performance and expandability. 

\subsubsection{Arguments Extraction}
\label{sec:scenario-understanding}
In order to understand the context and formalize the problem, the first step is to extract \textbf{the participants, positions, facts and rules expressed in natural language} from the passage and hypothesis of the question. 
An \textbf{assignment} represents a solution that assigns participants to positions, and has a group of values of three possible states: $(\text{\textit{True}}, \text{\textit{False}}, \text{\textit{Unknown}})$ representing whether a participant is assigned to a position.
The \textbf{rules} describe the constraints of assignments while the \textbf{facts} describe determined initial assignments explicitly mentioned in the context. 
We take the example in Figure \ref{fig:approach} as a running example to show the extracted participants, positions, facts and rules. 

Specifically, we extract the entities with a neural Named Entity Recognition (NER) model \cite{peters2017semi} and group the extracted entities into participants or positions. 
Rules and facts are identified by whether a sentence mentions determined assignment. 
We parse groups of entities that appear together in the leading sentence of the passage as groups of participants or positions, where participants always appear before positions. 
\subsubsection{Logical Function Extraction}
\label{sec:function-generation}
We introduce a set of predefined logical functions to express the constraints in the rules, which is the foundation of the reasoning process. 
A function consists of arguments and a executor, whose input is an assignment and the output is a \textit{Bool} value indicates whether the assignment satisfies the constraint.
The detailed definition of each function is listed in Appendix B. 
As the fragment shown in Table \ref{tab:functions}, the logical functions include following basic types:
\paragraph{Relational Function}
The relational functions, whose arguments involve participants or positions,  represent the constraints of the relationship between them.
For example, the function $\text{\textit{Before}}(Ginny, Fernando)$ indicates that \textit{Ginny} should be in the position before \textit{Fernando} in the ordering game. $\text{\textit{To}}(A,X)$ indicates that participant $A$ should be assigned to position $X$.
\paragraph{Compositional Function}
\begin{table*}[h]
\centering
\small
	\begin{tabular}{l|l|l|l}
		\hline
		\textbf{Type} & \textbf{Function} & \textbf{Args} & \textbf{Description} \\ \hline
		\multirow{3}{*}{\begin{tabular}[c]{@{}l@{}}Relational \\ Functions\end{tabular}} & \textit{Before/After} & \multirow{2}{*}{\begin{tabular}[c]{@{}l@{}}$participant_1$\\ $participant_2$\end{tabular}} & \begin{tabular}[c]{@{}l@{}}Whether $participant_1$ is in the \\ position before/after $participant_2$.\end{tabular} \\ \cline{2-2} \cline{4-4} 
		& \textit{Same/Different} &  & \begin{tabular}[c]{@{}l@{}}Whether $participant_1$ is in the \\ same/different position with $participant_2$.\end{tabular} \\ \cline{2-4} 
		& \textit{To} & \begin{tabular}[c]{@{}l@{}}$participant_1$\\ $position_1$\end{tabular} & \begin{tabular}[c]{@{}l@{}}Whether $participant_1$ is assigned\\ to $position_1$.\end{tabular} \\ \hline
		\begin{tabular}[c]{@{}l@{}}Compositional \\ Functions\end{tabular} & \textit{IfThen} & \begin{tabular}[c]{@{}l@{}}function set $F_1$\\ function set $F_2$\end{tabular} & \begin{tabular}[c]{@{}l@{}}If functions in $F_1$ satisfied,\\ then functions in  $F_2$ satisfied.\end{tabular} \\ \hline
		\begin{tabular}[c]{@{}l@{}}Counting \\ Functions\end{tabular} & \textit{FirstPos}/\textit{LastPos} & \begin{tabular}[c]{@{}l@{}}$participant_1$,\\ number $m$\end{tabular} & \begin{tabular}[c]{@{}l@{}}Whether $participant_1$ is assigned\\ to the first/last $m$ positions.\end{tabular} \\ \hline
	\end{tabular}
\caption{A fragment of the logical constraint function definition. }
\label{tab:functions}
\end{table*}
A compositional function expresses the relationship between two sets of functions, like the conditional rule (\textit{if-then} rule) and the \textit{if-and-only-if} rule. The arguments of compositional functions involve two sets of sub-functions.
For example, the rule ``\textit{If A serves on the X, then B serves on the Y.}" should be expressed as $\text{\textit{IfThen}}(\{To(A,X)\},\{To(B,Y)\})$.
\paragraph{Counting Function}
The counting functions focus on the calculation problem of participants under specific constraints. The arguments of counting functions involve a participant and a number. For example, $\text{\textit{LastPos}}(A,3)$ checks whether the participant A is assigned to the last 3 positions.

Based on the extracted arguments, we formalize the rules into logical functions. 
One straightforward way is to design a symbolic parsing method. 
For each function, we follow NSM \cite{liang2016neural} that uses trigger words to match a potential function. For example, the function \textit{Before} can be triggered by words ``\textit{before}" and ``\textit{earlier}". 
Then we select arguments (i.e., participants, positions, and numbers) based on their relative positions to the trigger word. 
The relational and counting functions can be constituted into compositional functions based on predefined grammar patterns. 
For example, for the grammar pattern ``\textit{If P, then Q}", Each function is grouped into the function set $F_1$ if it occurs in \textit{P}, or the function set $F_2$ if it occurs in $Q$. $F_1$ and $F_2$ are taken as the arguments of the function \textit{IfThen}. 

Furthermore, to handle the uncertain cases and improve the coverage of extracted functions, we build a neural semantic parsing model based on a pre-trained language model RoBERTa \cite{liu2019roberta}. 
It takes the sentence and two parsed arguments in the sentence as the input and predicts their potential function type. 
Specifically, given a rule as the input $X$, we follow \citet{xu2020syntax} and modify the input by adding special tokens ``@'' and ``\#'' before and after the first and second parsed arguments respectively. 
Then we encode sentence X with RoBERTa model as follows:
\begin{equation}
    H = \textit{RoBERTa}(X).
\end{equation}
Afterwards, we take the representation of the first ``@'' and ``\#'' for classification.
\begin{equation}
    function = argmax(\textit{classifier}([H^@;H^{\#}])),
\end{equation}
where [;] denotes concatenation, and the classifier is a linear layer followed by a softmax function.
, and $p$ is the possibilities distribution over class number.
Since there is no annotated data of corresponding logical functions, we need to construct the training data automatically.
The training data consist of 
(1) positive instances:
all the \textit{\{input: (rule, arguments); label: {function}\}} pairs that extracted by the symbolic parsing method from the training set; 
(2) negative instances: the same number of instances that have  arguments with no function related.

\subsubsection{Legitimate Assignments Deduction}
\label{sec:reasoning}
Given the extracted logical constraint functions and the initial assignment, we conduct reasoning to find the legitimate assignments that satisfy all the constraints.
\begin{figure}[h]
     \centering
     \includegraphics[width=0.45\textwidth]{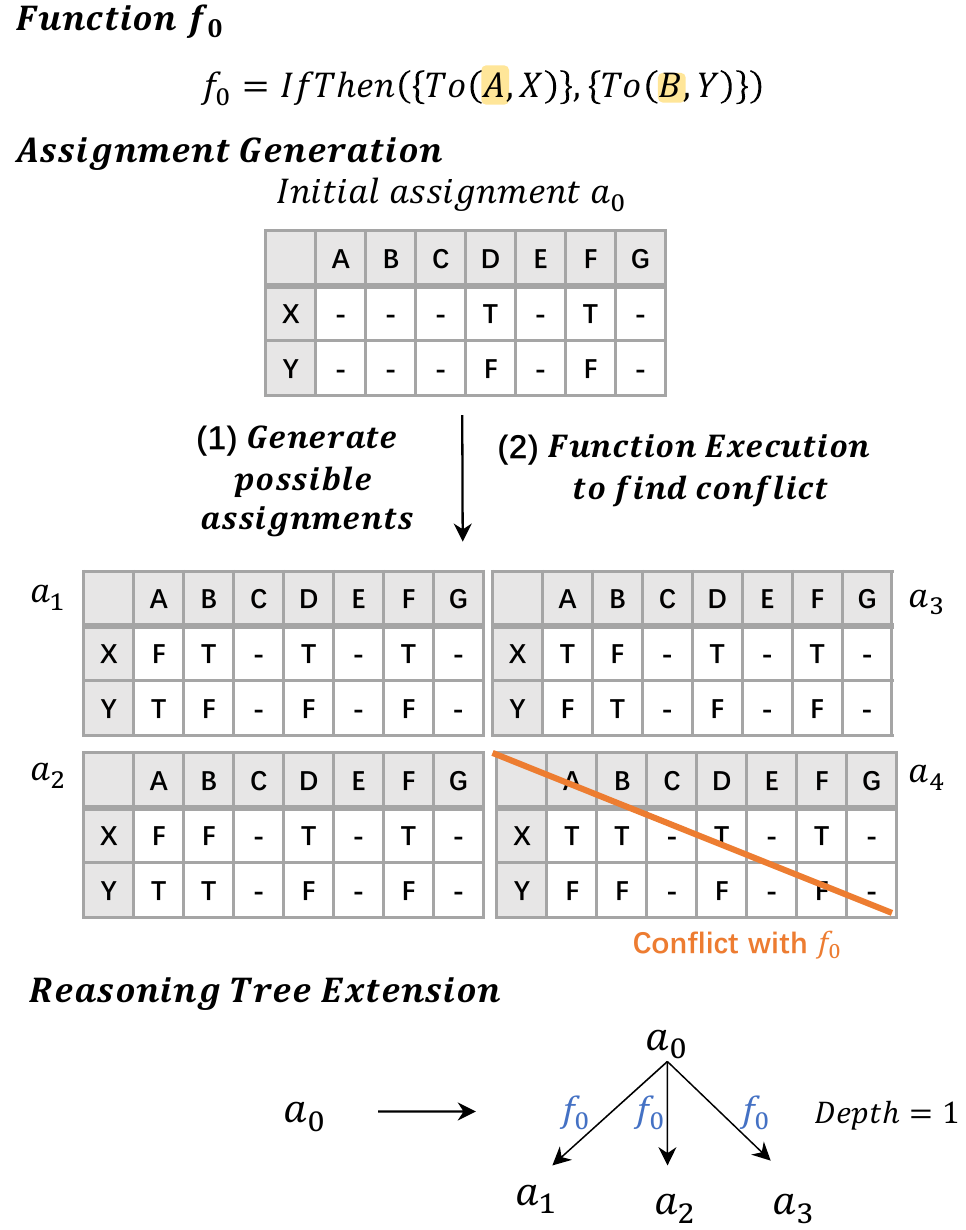}
     \caption{An example of the reasoning process. Newly added participants in $f_0$ are highlighted. (1) and (2) conducted recursively until $depth=n$. $(T/F/-) = (True/False/Unknown)$} 
     \label{fig:reasoning}
\end{figure}
The process is formulated into a tree-based reasoning algorithm. 
As shown in Figure \ref{fig:reasoning}, each node in a tree corresponds to an assignment and each edge indicates a logical function. A node $v$ with path $\{e_0,e_1,...,e_i\}$ from the root indicates that its assignment satisfies functions $\{f_0,f_1,...,f_i\}$. 
Suppose we have $n$ constraint functions, we need to find all the leaf nodes with depth $n$.
These leaf nodes satisfy all the functions and thus become legitimate assignments.

Therefore, we introduce how to construct the complete reasoning tree by the following steps:
\begin{itemize}
    \item[1)] Firstly, we start with the root, which is the certain initial assignment decided by facts. For the function $f_0$, we generate all possible assignments related to newly added arguments in $f_0$. As shown in the example in Figure \ref{fig:reasoning}, for the function $\text{\textit{IfThen}}({\text{\textit{To}}(A,X)},{\text{\textit{To}}(B,Y)})$, we generate all possible assignments related to the new participants $A$ and $B$.
    \item[2)] We execute $f_0$ to find all the legitimate assignments that satisfy $f_0$ as a group of children of the root. In the same example, we keep the assignments that meets $\text{\textit{IfThen}}({\text{\textit{To}}(A,X)},{\text{\textit{To}}(B,Y)})$.
    \item[3)] Then we select each child as a new root and select function $f_1$ for further extension of the reasoning tree.
\end{itemize}
 These processes are recursively conducted until depth $n$, which means that all the functions are used to construct the reasoning tree. 
 The tree-based manner reduces the computational complexity and can be further accelerated by ranking the functions.  The procedure is summarized into pseudo-code in Appendix A.
 Therefore, this algorithm has advantages of performing explicit interpretable reasoning over the extracted functions. 
 
\subsubsection{Answer Selection}
\label{sec:answer-selection}
Previous steps understand the passage and the question. 
In this part, we introduce how to analyze the options, and match the options with the deducted legitimate assignments beyond word-level for making a final prediction. 
Specifically, we can derive two types of information from an option:
\begin{itemize}
    \item[1)] \textbf{Assignment-based option} indicates an assignment. For example, ``\textit{A and C both serve on the X committee}" can be interpreted as: $\{(A,X)=\text{True}; (C,X)=\text{True}\}$. 
    For this type, we match the parsed option assignment with all the legitimate assignments and calculate an assignment-based matching score.
    \item[2)] \textbf{Function-based option} indicates an option representing a logical function, like ``\textit{The sedan is serviced earlier in the week than the roadster}", which can be parsed into the function ``\textit{Before(sedan, roadster)"}. 
    We execute the option-based function on the legitimate assignments to find the satisfiable option and calculate a function-based matching score.
\end{itemize}
These two types of scores are combined for making a conclusion. 
The question types and score calculating methods are summarized in the Appendix C.
\label{sec:length}
\section{Experiments}
In this section, we focus on evaluating the presented methods on AR-LSAT. We split the data into $\text{\textit{(train/dev./test)}} = (1,585/231/230)$. We also hold out a small test set for human evaluation. 
Moreover, case study illustrates the reasoning process of the ARM method by an explicit example. 
Lastly, we make error analysis to point out challenges in this task.
\subsection{Model Comparison}
\paragraph{Human Performance}
Since the dataset is based on a test designed for undergraduate students, we select nearly 100 instances in the AR-LSAT dataset and ask 10 undergraduate college students majoring in literature, commerce and law to answer these questions. In order to prevent the training bias, we select students who have not received LSAT professional training before.
We take their averaged performance as human performance and report it in Table \ref{table:model-comparison}. 
\paragraph{Transformer-based Methods}
We take various powerful Transformer-based pre-trained language models, including BERT \cite{devlin2018bert}, XLNet \cite{yang2019xlnet}, RoBERTa \cite{liu2019roberta}, and the recent ALBERT \cite{lan2019albert}), as the backbones of the Transformer-based methods and investigate their performance on the AR-LSAT dataset. 
 The implementation details of these models are given in Appendix D.
\begin{figure*}[tbp]
	\centering
	\includegraphics[width=\textwidth]{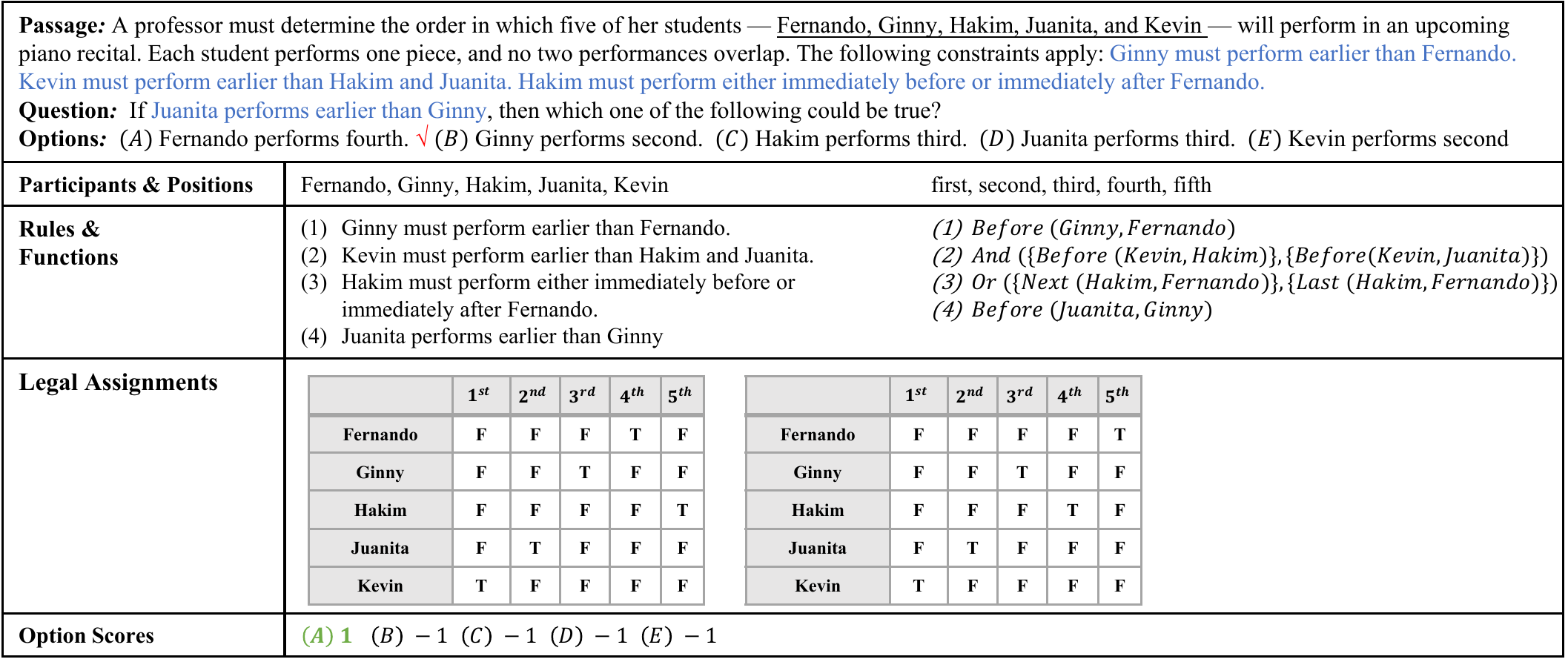}
	\caption{A case study on the AR-LSAT dataset. Our system correctly extracts participants, positions, and rules from the context. Afterwards, it interprets rules into logical functions. After deduction, our system finds legitimate assignments and makes the correct prediction. Rules are highlighted in blue.}
	\label{fig:case-study}
\end{figure*}

\paragraph{ARM}
To evaluate the performance of arguments extraction, we manually annotate the correct participants and positions in the development set as labels and report the accuracy and recall of in Table \ref{tab:condition-extraction}.
For function extraction, we define a API set to include roughly 20 types of logical functions like \textit{Before, After, To, IfThen} and realize their executors. The detailed definition of functions can be found in Appendix B. 
\begin{table}[h]
\centering
\begin{tabular}{lll}
\hline
             & Acc. (\%) & Recall (\%) \\ \hline
Participants & 96.17     & 92.88       \\
Positions    & 84.42     & 85.79       \\ \hline
\end{tabular}
\caption{Performance of extraction of participants and positions on the development set.}
\label{tab:condition-extraction}
\end{table}
\paragraph{Results}
\begin{table}[h]
\begin{tabular}{lcc}
\hline
Methods           & \begin{tabular}[c]{@{}l@{}}Dev. \\ Acc (\%)\end{tabular} & \begin{tabular}[c]{@{}l@{}}Test\\ Acc (\%)\end{tabular} \\ \hline
Human Performance &  -                                                &  59.7\%                                                  \\ \hline
Random Guess      & 20.0\%                                                  & 20.0\%                                                  \\ \hline
BERT              & 23.4\%                                                 & 21.4\%                                                        \\
XLNet             & 23.8\%                                                   & 22.5\%                                                        \\
RoBERTa           & 24.2\%                                                   &       23.1\%                                                   \\ 
ALBERT             & 24.4\%                                                   & 23.0\%                                                        \\
\hline
ARM       & 34.2\%                                                  & 30.9\%                                                  \\ \hline
\end{tabular}
\caption{
The performance on the AR-LSAT dataset. }
\label{table:model-comparison}
\end{table}
In Table \ref{table:model-comparison}, we report the performance of different methods and human performance on the development and test set. 
Firstly, we observe that the Transformer-based models struggle to do well on this task, and achieve close performance with random guess. This observation indicates that analytical reasoning is extremely challenging for current neural pre-trained language models as it requires the ability of complex reasoning.
In addition, ARM with context understanding and explicit reasoning process outperforms Transformer-based method with 34.2\% accuracy on the development set and 30.9\% accuracy on the test set. 
It is also noticed that the performance of both our system and baselines are still far from human performance, leaving significant opportunities for further exploration. 

\subsection{Case Study}
We present a case study in Figure \ref{fig:case-study} to illustrate the reasoning process of the ARM framework with interpretable results. 
ARM extracts correct arguments from the context, and interprets the rules into logical constraint functions. 
Afterwards, it performs deduction to find legitimate solutions. 
Lastly, it matches the options with the legitimate solutions and calculates a score for each option. 
Option $A$ achieves the highest score because it accords with legitimate assignments. 
This analysis demonstrates that ARM has better explicit interpretable reasoning ability.

\subsection{Error Analysis}
We randomly select 50 instances that are wrongly predicted by ARM from the development set and manually summarize the major error types. 

The dominant error type is that some rules with complex semantics are not covered by current constraint logical function set. For example, given a rule
``\textit{Each crew member does at least one task during the installation.}" , we should map ``\textit{At least}" to function \textit{AtLeastNum}.

The second type of errors is caused by failing to extract correct participants or positions by the NER model and predefined matching pattern. 

The third error type is caused by the lack of basic commonsense knowledge, which is required for understanding the concept in the rules. For example, when a passage mentioned ``\textit{Six entertainers should be scheduled at 9:00 A.M., 2:00 P.M., etc}" and the rule is ``\textit{Some participants should be scheduled in the morning.}", the system fails to match the \textit{morning} with a specific time zone.
\subsection{Discussion}
We would like to further highlight important directions to facilitate research on analytical reasoning.

One of the major challenges lies in deep understanding of the knowledge in the context, like parsing the rules into logically equivalent symbolic functions. 
Deriving machine-understandable functions from natural language is an essential step towards deeper understanding and reasoning.
Although supervised semantic parsing has achieved promising progress in recent years, obtaining complete human-annotated logical functions is impractical for this task.
Therefore, further study can focus on function extraction with no annotated functions or small amount of annotated functions.

Furthermore, a better inference engine built upon logical functions is also essential because AR questions require deeper reasoning abilities far beyond just understanding the literal clues. 
Standard symbolic systems like expert systems can provide explicit reasoning, but they are difficult to deal with uncertainty in data. 
Although neural-based methods are more flexible at dealing with uncertainty, they still struggle to perform interpretable and explicit reasoning. 
It is promising to better integrate neural and symbolic systems to improve this task with deeper reasoning ability.  
\section{Conclusion}
In this paper, we study the challenging task of analytical reasoning and introduce a dataset AR-LSAT to facilitate research on analytical reasoning.
We analyze the knowledge understanding and reasoning ability required for this task and present
two basic approaches: a Transformer-based approach and a logical-level reasoning framework, named Analytical Reasoning Machine (ARM). 
ARM extracts symbolic knowledge, including participants, facts and rules mentioned in the context and extract logical functions from the rules. 
Afterwards, it performs deep reasoning to find all the legitimate solutions to the problem posed and finally makes a prediction.
ARM sheds a light on the reasoning procedure for analytical reasoning, and each component can be further developed. 
Experiments show that this task is very challenging for current Transformer-based pre-trained language models and ARM outperforms them with better performance and interpretability. Further discussions are made to shed light on important future directions.
\bibliography{anthology,custom}
\bibliographystyle{acl_natbib}

\appendix
\newpage
\section{Pseudo-code of Legitimate Assignments Deduction}

\begin{algorithm}[h]
\centering
\footnotesize
\begin{algorithmic}[1]
\Require
A set of constraint functions $F=\{f_0,f_1,...,f_n\}$ and an initial assignment $a_0$
\Function{ConstructTree}{node,functions,depth,n}
    \If{depth == $n$}:
    \State
        \Return
    \EndIf
    \State
    function = functions[depth]
    \State
    old\_pars = node.participants
    \State
    old\_assign = node.assignment
    \State
    new\_pars = find\_new\_participant(function, old\_pars)
    \State
    all\_assign = gen\_all\_assign(old\_assign, new\_pars)
    \State
    satisfied = find\_satisfied(all\_assign, function)
    \State
    depth = depth+1
    \State
    children = update\_notes(node, satisfied, new\_pars)
    \For{child in children}
    \State
    \Call{ConstructTree}{child, functions, depth, $n$}
    \EndFor

\EndFunction
\State
root = Node($a_0$)
\State
depth = 0
\State
$n$ = length of $F$
\State
complete\_tree = \Call{ConstructTree}{root, $F$, depth, $n$}
\State
legitimate = nodes in complete\_tree with depth $n$
\State
\Return legitimate
\end{algorithmic}
\end{algorithm}
\section{Function Definition}
In this part, we present the detailed description and trigger words for each logical constraint functions in Table \ref{tab:function-definition}.
\section{Question Type}
\label{appendix:ques-type}
\begin{table*}[ht]
\small
\begin{tabular}{ll}
\hline
\textbf{Question Type} & \textbf{Example} \\ \hline
Acceptable solution & Which one of the following could be the schedule of the students' reports? \\ \hline
Complete list & \begin{tabular}[c]{@{}l@{}}Which one of the following could be a complete and accurate list of \\ the books placed on the bottom shelf?\end{tabular} \\ \hline
Could be true/false with condition & If Himalayans are not featured on day 7. which one of the following could be true? \\ \hline
Must be true/false with condition & If Theresa tests G on the second day. then which one of the following must be true? \\ \hline
Negation & P CANNOT be performed at? \\ \hline
Substitution & \begin{tabular}[c]{@{}l@{}}Which one of the following. if substituted for the condition that Waite's audition\\  must take place earlier than the two recorded auditions. \\ would have the same effect in determining the order of the auditions?\end{tabular} \\ \hline
Condition for unique solution & \begin{tabular}[c]{@{}l@{}}The assignment of parking spaces to each of the new employees is fully and uniquely\\ determined if which one of the following is true?\end{tabular} \\ \hline
Calculation & How many of the students are there who could be the one assigned to 1921? \\ \hline
Earliest/latest position & \begin{tabular}[c]{@{}l@{}}If Zircon performs in an earlier slot than Yardsign. which one of the following \\ is the earliest slot in which Wellspring could perform?\end{tabular} \\ \hline
Maximum/minimum members & What is the minimum number of solos in which Wayne performs a traditional piece? \\ \hline
\end{tabular}
\caption{Question types of AR-LSAT dataset.}
\label{tab:question-type-example}
\end{table*}

In this part, we list common question types in the AR-LSAT datasets and give examples in Table \ref{tab:question-type-example}.
We further introduce how we calculate a score for dominant question type with a group of legitimate assignments.
\begin{itemize}
    \item[1)] \textbf{Must be true/false}: this question type needs to select answer that must be true in all the assignments. 
    We match all the assignments with the option. If one option accords/conflicts with one assignment, the single matching score will be 1/-1, otherwise the score will be 0. 
    We then calculate the sum of all the matching scores as the final score.
    \item[2)] \textbf{Could be true/false}: this question type needs to select answer that could be true in one of the legitimate assignments. 
    We match all the assignments with the option. If one option accords/conflicts with one assignment, the single matching score will be 1/-1, otherwise the score will be 0. 
    We then calculate the maximum matching scores as the final score. The \textit{Acceptable solution} question type also use this method to calculate score.
    \item[3)] \textbf{Maximum number of participants in a position}: this question type needs to calculate the maximum possible number of participants in a specified position (group). We calculate the maximum number of participants in all the legetimate assignments and calculate the absolute difference with the number in the option as the final score.
    \item[4)] \textbf{Find the earliest position of a participant}: this question type needs to calculate the earliest possible position of a specific participant. We calculate the index of the earliest position of the participant in all the legitimate assignments and calculate the absolute difference with the number in the option as the final score.
    \item[5)] \textbf{Count the number of possible positions that a participant can be assigned in}: for this question type, we count all the non-repetitive assignments of the specific participant and calculate the absolute difference with the number in the option as the final score.
\end{itemize}
\section{Baseline Models}
\subsection{Descriptions}
\begin{itemize}
    \item \textbf{LSTM} \cite{gers1999learning} is a classical RNN-based model. We apply Bi-LSTM with GloVE \cite{pennington-etal-2014-glove} embedding.
    \item \textbf{BERT} \cite{devlin2018bert} is a transformer-based model pre-trained on BooksCorpus and Wikipedia with two unsupervised learning task: Masked LM and Nest Sentence Prediction. 
    \item \textbf{XLNet} \cite{yang2019xlnet} is also a transformer-based model, pre-trained on BooksCorpus, Wikipedia, Giga5, ClueWeb 2012-B and Common Crawl with Permutation Language Modeling. 
    \item \textbf{RoBERTa} \cite{liu2019roberta} is a transformer-based model with the same model structure as BERT but trained on a larger corpus and on a different training setting.
    \item \textbf{ALBERT} \cite{lan2019albert} is a most recent transformer-based pre-trained model. ALBERT uses parameter-reduction techniques that support large-scale configurations.
\end{itemize}
\subsection{Implementation Details}
For all the baselines, we employ cross-entropy loss as the loss function and select AdamW as the optimizer for model training/ fine-tuning. 
These baselines add a simple classification layer on the top of them and take the the last hidden state as the input. For all the Transformer-based models, we employ base model as the backbone. 

\begin{table*}[h]
\resizebox{\textwidth}{!}{%
\begin{tabular}{|l|l|l|l|l|}
\hline
Type                                                                             & Function    & Arguments                                                                                & Description                                                                                                               & Trigger Words                                                                 \\ \hline
\multirow{10}{*}{\begin{tabular}[c]{@{}l@{}}Relational\\ Functions\end{tabular}} & Before      & \multirow{9}{*}{\begin{tabular}[c]{@{}l@{}}participant 1\\ participant 2\end{tabular}}   & \begin{tabular}[c]{@{}l@{}}whether participant 1 is in the \\ position before  participant 2\end{tabular}                 & \begin{tabular}[c]{@{}l@{}}before, above,\\ precede, earlier\end{tabular}     \\ \cline{2-2} \cline{4-5} 
                                                                                 & After       &                                                                                          & \begin{tabular}[c]{@{}l@{}}whether participant 1 is in the \\ position after participant 2\end{tabular}                   & \begin{tabular}[c]{@{}l@{}}after, larger, higher\\ bigger, older\end{tabular} \\ \cline{2-2} \cline{4-5} 
                                                                                 & Last        &                                                                                          & \begin{tabular}[c]{@{}l@{}}whether participant 1 is in the \\ last position of  participant 2\end{tabular}                & \begin{tabular}[c]{@{}l@{}}immediately before,\\ last\end{tabular}            \\ \cline{2-2} \cline{4-5} 
                                                                                 & Next        &                                                                                          & \begin{tabular}[c]{@{}l@{}}whether participant 1 is next \\ to participant 2\end{tabular}                                 & \begin{tabular}[c]{@{}l@{}}immediately after,\\ next\end{tabular}             \\ \cline{2-2} \cline{4-5} 
                                                                                 & Adjacent    &                                                                                          & \begin{tabular}[c]{@{}l@{}}whether participant 1 is\\ neighboring to participant 2\end{tabular}                           & \begin{tabular}[c]{@{}l@{}}neighboring,\\ adjacent\end{tabular}               \\ \cline{2-2} \cline{4-5} 
                                                                                 & Different   &                                                                                          & \begin{tabular}[c]{@{}l@{}}whether participant 1 in the different \\ position with participant 2\end{tabular}             & different                                                                     \\ \cline{2-2} \cline{4-5} 
                                                                                 & Same        &                                                                                          & \begin{tabular}[c]{@{}l@{}}whether the first participant in the same \\ position with the second participant\end{tabular} & same, also                                                                    \\ \cline{2-2} \cline{4-5} 
                                                                                 & BeforeEqual &                                                                                          & \begin{tabular}[c]{@{}l@{}}whether participant 1 before \\ or equals to the position of participant 2\end{tabular}        & no later                                                                      \\ \cline{2-2} \cline{4-5} 
                                                                                 & AfterEqual  &                                                                                          & \begin{tabular}[c]{@{}l@{}}whether participant 1 after or equals \\ to the position of  participant 2\end{tabular}        & no earlier                                                                    \\ \cline{2-5} 
                                                                                 & To          & \begin{tabular}[c]{@{}l@{}}participant\\ position\end{tabular}                           & \begin{tabular}[c]{@{}l@{}}Whether the participant is \\ assigned to the position\end{tabular}                            & to, on, give, in                                                              \\ \hline
\multirow{6}{*}{\begin{tabular}[c]{@{}l@{}}Compos.\\ Functions\end{tabular}}     & IfThen      & \multirow{6}{*}{\begin{tabular}[c]{@{}l@{}}function set 1\\ function set 2\end{tabular}} & \begin{tabular}[c]{@{}l@{}}If rules in rule set 1 satisfied,\\ then rules in rule set 2 satisfied\end{tabular}            & If... then, If ... , ...                                                      \\ \cline{2-2} \cline{4-5} 
                                                                                 & IFF         &                                                                                          & \begin{tabular}[c]{@{}l@{}}Rules in rule set 1 satisfied if and \\ only if rules in rule set 2 satisfied\end{tabular}     & if and only if                                                                \\ \cline{2-2} \cline{4-5} 
                                                                                 & And         &                                                                                          & \begin{tabular}[c]{@{}l@{}}Rules in rule set 1 satisfied and\\ rules in the rule set  2 satisfied\end{tabular}            & and                                                                           \\ \cline{2-2} \cline{4-5} 
                                                                                 & Or          &                                                                                          & \begin{tabular}[c]{@{}l@{}}Rules in rule set 1 satisfied or\\ rules in rule set 2 satisfied\end{tabular}                  & or                                                                            \\ \cline{2-2} \cline{4-5} 
                                                                                 & Unless      &                                                                                          & \begin{tabular}[c]{@{}l@{}}Rules in rule set 1 satisfied unless\\ rules in rule set 2 satisfied\end{tabular}              & unless                                                                        \\ \cline{2-2} \cline{4-5} 
                                                                                 & Neither     &                                                                                          & \begin{tabular}[c]{@{}l@{}}Neither rules in rule set 1 satisfied \\ nor rules in rule set  2 satisfied\end{tabular}       & Neither ... nor ...                                                           \\ \hline
\multirow{2}{*}{\begin{tabular}[c]{@{}l@{}}Counting \\ Functions\end{tabular}}   & FirstPos    & \multirow{2}{*}{\begin{tabular}[c]{@{}l@{}}participant\\ number\end{tabular}}            & \begin{tabular}[c]{@{}l@{}}Whether the participant is in the \\ last (number) positions\end{tabular}                      & \begin{tabular}[c]{@{}l@{}}one of the \\ last (number)\end{tabular}           \\ \cline{2-2} \cline{4-5} 
                                                                                 & LastPos     &                                                                                          & \begin{tabular}[c]{@{}l@{}}Whether the participant is in the \\ first (number) positions\end{tabular}                     & \begin{tabular}[c]{@{}l@{}}one of the \\ first (number)\end{tabular}          \\ \hline
\end{tabular}%
}
\caption{Detailed function descriptions and corresponding trigger words}
\label{tab:function-definition}
\end{table*}

\end{document}